\documentclass[conference]{IEEEtran}
\IEEEoverridecommandlockouts
\usepackage{cite}
\usepackage{amsmath,amssymb,amsfonts}
\usepackage{algorithmic}
\usepackage{graphicx}
\usepackage{textcomp}
\usepackage{xcolor}

\usepackage{xspace}
\usepackage{multirow}
\usepackage{booktabs}
\newcommand{\model}{ClearGAE\xspace}

\usepackage{colortbl}
\usepackage{threeparttable}
\definecolor{mygray}{gray}{0.925}

\usepackage{amsmath,amsfonts,bm}

\def\eqref#1{equation~\ref{#1}}

\def\1{\bm{1}}

\DeclareMathAlphabet{\mathsfit}{\encodingdefault}{\sfdefault}{m}{sl}
\SetMathAlphabet{\mathsfit}{bold}{\encodingdefault}{\sfdefault}{bx}{n}

\def\gA{{\mathcal{A}}}

\def\gE{{\mathcal{E}}}

\def\gG{{\mathcal{G}}}
\def\gH{{\mathcal{H}}}

\def\gL{{\mathcal{L}}}

\def\gN{{\mathcal{N}}}

\def\gV{{\mathcal{V}}}

\def\gX{{\mathcal{X}}}

\def\BibTeX{{\rm B\kern-.05em{\sc i\kern-.025em b}\kern-.08em
    T\kern-.1667em\lower.7ex\hbox{E}\kern-.125emX}}
\begin{document}

\title{Preserving Node Distinctness in Graph Autoencoders via Similarity Distillation}

\author{\IEEEauthorblockN{Ge Chen\IEEEauthorrefmark{1}\IEEEauthorrefmark{2}\thanks{\IEEEauthorrefmark{1}Equal contribution},
Yulan Hu\IEEEauthorrefmark{1}\IEEEauthorrefmark{3},
Sheng Ouyang\IEEEauthorrefmark{3}, 
Yong Liu\IEEEauthorrefmark{3} and
Cuicui Luo\IEEEauthorrefmark{2}}
\IEEEauthorblockA{\IEEEauthorrefmark{2}University of Chinese Academy of Sciences, Beijing, China}
\IEEEauthorblockA{\IEEEauthorrefmark{3}Renmin University of China, Beijing, China}
\IEEEauthorblockA{\{chenge221, luocuicui\}@mails.ucas.ac.cn \\
\{huyulan, ouyangsheng, liuyonggsai\}@ruc.edu.cn}
}

\maketitle
\newcommand{\vpara}[1]{\vspace{0.04in}\noindent\textbf{#1}\xspace}
\begin{abstract}
Graph autoencoders (GAEs), as a kind of generative self-supervised learning approach, have shown great potential in recent years. GAEs typically rely on distance-based criteria, such as mean-square-error (MSE), to reconstruct the input graph. However, relying solely on a single reconstruction criterion may lead to a loss of distinctiveness in the reconstructed graph, causing nodes to collapse into similar representations and resulting in sub-optimal performance. To address this issue, we have developed a simple yet effective strategy to preserve the necessary distinctness in the reconstructed graph. Inspired by the knowledge distillation technique, we found that the dual encoder-decoder architecture of GAEs can be viewed as a teacher-student relationship. Therefore, we propose transferring the knowledge of distinctness from the raw graph to the reconstructed graph, achieved through a simple KL constraint. Specifically, we compute pairwise node similarity scores in the raw graph and reconstructed graph. During the training process, the KL constraint is optimized alongside the reconstruction criterion. We conducted extensive experiments across three types of graph tasks, demonstrating the effectiveness and generality of our strategy. This indicates that the proposed approach can be employed as a plug-and-play method to avoid vague reconstructions and enhance overall performance. 
\end{abstract}

\section{Introduction}~\label{sec:intro}
Generative graph self-supervised learning (SSL), exemplified by Graph Autoencoders (GAEs) \cite{gae,graphmae,union_model_hu}, has garnered significant research interest in recent years. As a subset of SSL \cite{ssl_taxonomy}, generative SSL enables the model to generate its own supervisory signals, which subsequently enhance performance during the training process.

Existing GAEs typically rely on a single graph characteristic for reconstruction, such as graph features or structures. For example, the masked GAE, GraphMAE~\cite{graphmae}, reconstructs the raw input graph features using decoded representations from the GNN decoder. In its companion work, GraphMAE2~\cite{graphmae2}, reconstruction is performed using the hidden representations from the encoder, thereby developing multiple reconstruction channels for robust training. The most recent work, GA$^2$E~\cite{union_model_hu}, aims to unify diverse graph tasks within the GAE framework and also employs dual features as the training objective.
The feature reconstruction on graph essentially resemble the image reconstruction in CV domain, with the latter focus on pixel-level feature. In addition, there exists substantial work that adopts structure reconstruction as an objective. GAE/VGAE~\cite{gae} can be considered early work of GAEs, focusing on reconstructing the adjacency matrix to facilitate downstream link prediction tasks, which is strongly relevant. Unlike GA$^2$E~\cite{union_model_hu}, which serves as a preliminary attempt to unify diverse graph tasks, S2GAE~\cite{S2GAE} uses edge reconstruction as the training criterion, yielding impressive performance. Additionally, some GAEs explore specific graph applications, such as heterogeneous graph learning~\cite{hgmae,hgvae} and imbalanced node classification~\cite{vigraph}.

\begin{figure}[h]
\centering 
\includegraphics[width=0.4\textwidth]{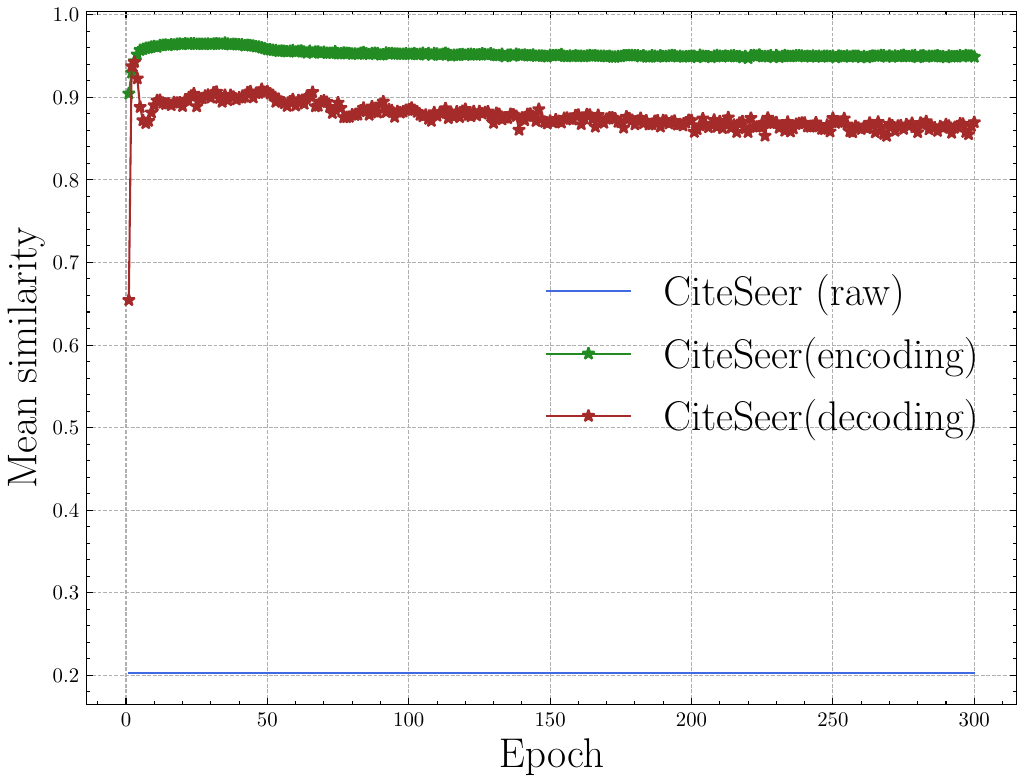}
\caption{The pairwise nodes similarity evolving at encoding and decoding stage.} 
\label{fig:similary_decode}
\end{figure}

Despite the progress made by the existing GAEs, a critical issue still exists. Conventional GAEs typically employ Mean Squared Error (MSE) \cite{mse} or the scaled cosine error proposed by GraphMAE \cite{graphmae} as the reconstruction criterion. These two functions are used to measure the discrepancy between two points, i.e., the node or edge representation in the input graph and the reconstructed output, respectively. As the common reconstruction paradigm employed by almost all GAEs, the reconstruction criteria actually neglect a significant issue: \textbf{the MSE or SCE criterion optimizes on the point-to-point instances, while losing the pairwise nodes' distinctness during the reconstruction process.} \cite{extracting_high_frequency} analyzed and revealed that under a dual GNN-MLP distillation framework, the loss of high-frequency knowledge causes the MLP student to acquire smooth knowledge that lacks distinctness. We conducted an analysis experiment on the Citeseer dataset to learn the pairwise nodes' similarity evolution for the reconstructed nodes during the whole training process. For each node, we calculated the average similarity of the node with its neighbors and presented the result in Figure \ref{fig:similary_decode}. It can be seen that the similarity score quickly rises to a high level at the beginning of the training, and remains relatively stable throughout the whole training stage. In comparison, the pairwise nodes' similarity in the raw graph is low. To illustrate, the oversmoothing problem \cite{oversmooth}, where nodes become increasingly similar after stacking multiple GNN layers, is a known issue in graph learning. A common practice to avoid this is to use fewer layers. However, within the GAE framework, a similar phenomenon occurs regardless of the number of layers employed.

To address this issue, we develop a simple yet effective strategy to preserve necessary node similarities, inspired by knowledge distillation on graphs \cite{gnn_knowledge_distill}. As nodes on the reconstructed graph lack the essential distinctness of those on the raw graph, we employ a knowledge distillation technique to transfer knowledge from the raw graph to the reconstructed graph. Specifically, for each node, we compute the similarity between it and its first-order neighbors and treat this knowledge as the teacher. Similarly, we compute the similarity for each node on the reconstructed graph. Using the two sets of similarity scores, we apply a Kullback-Leibler (KL) divergence constraint to force the student to imitate the teacher. This simple strategy effectively preserves the necessary distinctness between nodes, prompting the GAE to update with a clearer, more accurate decoded graph, thereby enhancing overall model performance, we name the proposed method as \model. To summarize, our contributions are as follows:
\begin{itemize}
    \item For the first time, we identify the problem in GAEs: reliance on a single reconstruction criterion may lose the necessary distinctness between nodes, leading to sub-optimal results.
    \item We propose a simple yet efficient strategy inspired by knowledge distillation to address the issue. This plug-and-play solution can be seamlessly integrated into GAEs.
    \item We conduct comprehensive experiments across three kinds of graph tasks. The results demonstrate the effectiveness and generality of the proposed method.
\end{itemize}

\section{Related Works}
\subsection{Generative Self-Supervised Graph Learning}\label{A}
GraphMAE/GraphMAE2~\cite{graphmae2}, MaskGAE~\cite{maskedgae}/HGMAE~\cite{hgmae}, Bandana~\cite{bandana} mask and reconstruct the node features, edges/paths, a portion of information through each edge. SimSGT~\cite{simsgt} suggests that the reconstruction of subgraph-level information improves the efficacy of molecular representation learning. Existing researches concentrate on the similarity between reconstructed node features and their original counterparts, while ignore the differences between paired nodes after reconstruction. However, these differences are also crucial for the original node.

Generative self-supervised learning (SSL) aims to generate new graphs that closely resemble the given input. In our study, we concentrate on the graph autoencoding family (GAEs) of generative SSL. These GAEs typically involve encoding the inputs into a latent feature space and reconstructing the graphs from these encoded representations. The pioneering work, known as GAE/VGAE \cite{gae}, leverages a GNN encoder and a dot-product decoder to reconstruct the structural information. For different downstream tasks, later GAEs explore the recovery of various kinds of graph information, which includes features \cite{gaae}, structure \cite{gae}, or a combination of both \cite{dge}. Meanwhile, several works focus on attaining an improved latent space to generate new representations. Since the simplistic prior assumption leads to a suboptimal latent space, SIG-VAE \cite{sig-vae} seeks to enhance VGAE by introducing Semi-Implicit Variational Inference. ARGA/ARVGA \cite{arga} integrates an adversarial training module into the GAE/VGAE framework, aligning the latent space with the prior distribution seen in actual data.

However, early GAEs without masking performed unsatisfactorily in the classification task \cite{graphmae}. Recently, the outstanding performance of GraphMAE \cite{graphmae}, on par with contrastive learning methods, has spotlighted the “mask-then-reconstruct” scheme in the graph domain. Subsequent GraphMAE series have pivoted towards graph masking modeling and the reconstruction of the graph. GraphMAE~\cite{graphmae}, GraphMAE2 \cite{graphmae2}, MaskGAE \cite{maskedgae}, HGMAE \cite{hgmae}, and Bandana \cite{bandana} mask and reconstruct various components: node features, edges/paths, and portions of information through each edge, respectively. SimSGT \cite{simsgt} suggests that the reconstruction of subgraph-level information improves the efficacy of molecular representation learning. Existing research concentrates on the similarity between reconstructed node features and their original counterparts while ignoring the differences between paired nodes after reconstruction. However, these differences are also crucial for the integrity of the original node.

\section{Methodology}
\subsection{Preliminaries}\label{model:pre}
\vpara{Notations.} Given a graph $\gG = (\gV, \gE, \gX, \gA)$, where $\gV$ is the set of $N$ nodes and $\gE \subseteq \gV \times \gV$ is the set of edges. $\gX \in \mathbb{R}^{N \times F_s}$ is the input node feature, $F_s$ denotes the feature dimension. Each node $v \in \gV$ is associated with a feature vector $x_v \in \gX$, and each edge $e_{u,v} \in \gE$ denotes a connection between node $u$ and node $v$. The graph structure can also be represented by an adjacency matrix $\gA \in \{0,1\}^{N \times N}$ with $A_{u,v} = 1$ if $e_{u,v} \in \gE$, and $A_{u,v} = 0$ if $e_{u,v} \notin \gE$. For masked graph autoencoder, we use $\lambda$ to denote the mask ratio. The encoder $f_E$ maps the input masked graph into latent representation, then a decoder $f_D$ endeavour to recover the input graph from the latent representation.

\subsection{The losing of pairwise nodes distinctness}
\begin{figure}[h]
\centering 
\includegraphics[width=0.4\textwidth]{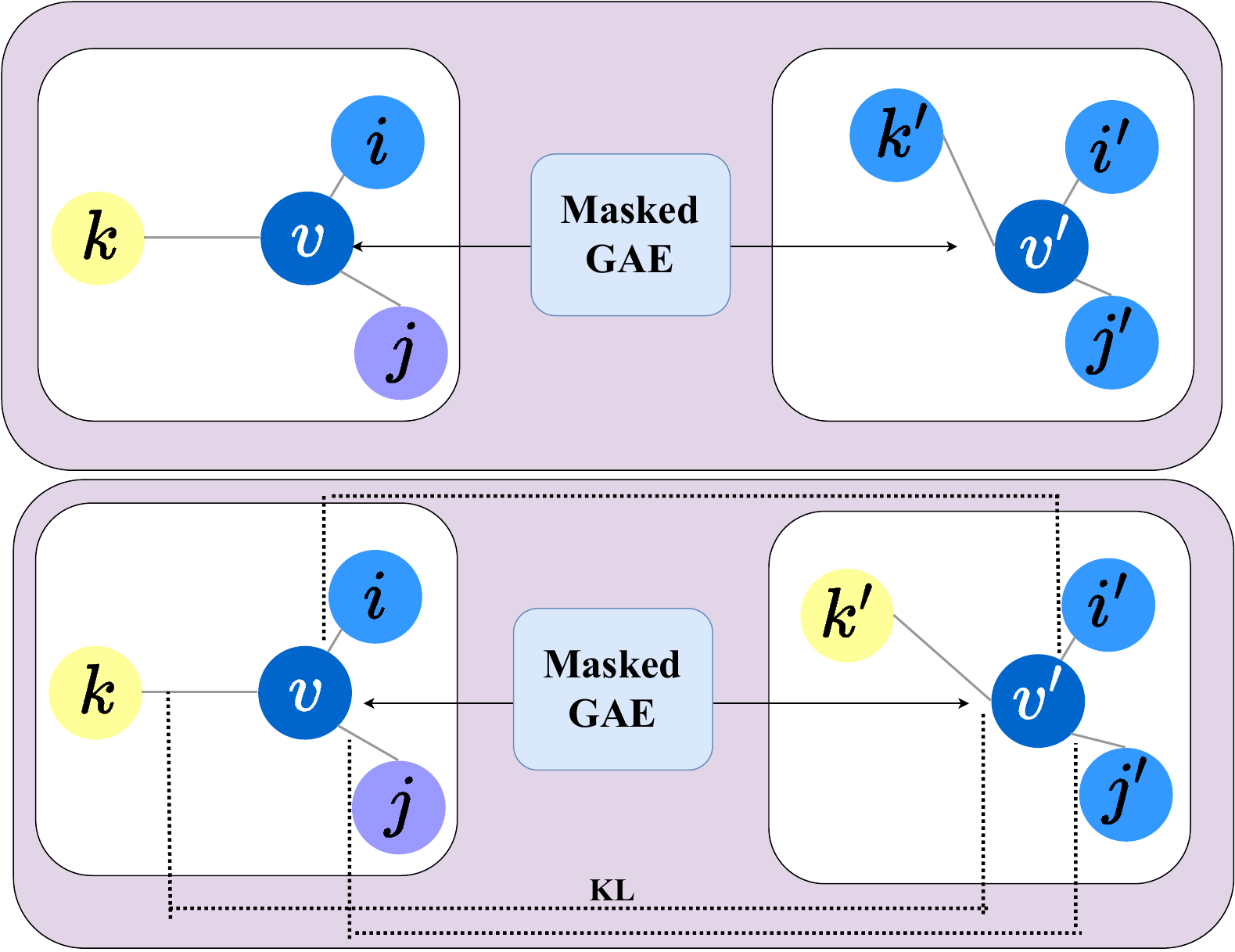}
\caption{Nodes similarity changes at decoding.} 
\label{fig:similary_evolving}
\end{figure}

We conduct preliminary experiments in Section \ref{sec:intro}, showing the change in pairwise node similarity during the training process. Specifically, for each node ( v ) and its neighbor nodes set ( $\gN_v$ ), we measure its average similarity with its corresponding neighbors: 
\begin{equation} 
\mathcal{S}_v = \frac{1}{|\mathcal{N}v|} \sum{i \in \mathcal{N}_v} \text{sim}(v, i), \label{equa:average_similarity_score}
\end{equation}
where $\mathcal{S}$ denotes the similarity score matrix for the nodes on the graph.

In this subsection, we present an example to illustrate the phenomenon. As shown in Figure \ref{fig:similary_evolving}, a center node ( v ) is connected with three adjacent nodes: ( i ), ( j ), and ( k ). The standard GAEs perform point-wise feature reconstruction, acquiring new representations for nodes ( v ), ( i ), ( j ), and ( k ). These nodes become similar after reconstruction. To address this issue, we propose a strategy that adds constraints on the similarity of pairwise nodes, which we will introduce next.

\subsection{Constraint on pairwise similarity}
As analyzed above, the feature reconstruction is prone to lead to blur decoding result, with the distinct information between nodes getting drowned as the reconstruction proceeds. This phenomenon is infeasible for GAE learning, the similarity growing between the pairwise nodes indicates the holistic nodes set collapse to identical. To address this issue, we propose a intuitive strategy to preserve the necessary discrepancy among nodes. As Figure~\ref{fig:similary_evolving} illustrates, the nodes in the encoder exhibit distinctness among each other, the adjacent nodes around node $v$ showcasing diverse impact on $v$, the similarity scores between each adjacent node and center node $v$ are different too. However, during the decoding stage, the impact of different adjacent nodes on $v$ is growing identical, indicating those nodes are getting more semantic similar as the training proceeding. Consider the image reconstruction in the CV domain, the convergent of pixel-level feature means the image losing its differentiation, making the reconstructed image looks blur. Since we focus on feature reconstruction, we consider the phenomenon occurs in CV domain also exists in the GAE learning. 

To address this problem, we develop a simple yet effective strategy to facilitate reconstructing with appropriate constraint. To encode the graph,  we first mask the input graph to get a corrupted graph, denoted as $\tilde{\gG}={\tilde{\gX}, \gA}$. Then, we use a single-layer GNN as the encoder to map $\tilde{\gG}$ into latent representation, acquiring $\gH$. Based on $\gH$, another decoder is employed to recover the feature from $\gH$, acquiring the reconstructed graph feature $\hat{\gX}$. The conventional GAEs typically employ MSE or SCE~\cite{graphmae,graphmae2} as the reconstruction criterion, which is performed on the masked feature, formed as:
\begin{equation}
    \gL_{rec} = \text{Rec}(\gX_m, \hat{\gX}_m).
\end{equation}

Beyond the reconstruction criterion, we aim to preserve the distinctness between nodes. We achieve this by adding restriction on the pairwise nodes similarity during decoding stage, which can be seen as a kind of knowledge distillation from the raw graph to the decoded graph. Specifically, the proposed strategy contains three steps. \textbf{First}, for each node $v$, we calculate its similarity with its first-order neighbors $\gN_v$, acquiring the similarity score set $\mathcal{S}_v = \left\{ \text{Sim}(v, i) , i \in \mathcal{N}_v  \right\}
$. $\mathcal{S}_v$ characterize the local semantic similarity between each pairwise nodes set. \textbf{Second}, for each node $v^{'}$ at the decoding phase, we follow the first step to calculate the similarity between $v^{'}$ and its first-order neighbors, acquiring the similarity score set $\mathcal{S^{'}}_v = \left\{ \text{Sim}(v^{'}, i) , i \in \mathcal{N}_v \right\}
$. \textbf{Third}, as analyzed above, the nodes at the decoding phase lacks differentiation. To avoid this, we treat the similarity score set in raw graph as teacher, and distill the distinctness to the pairwise nodes at decoding phase. We employs the KL divergence as the constraint to prevent the nodes at decoding phase collapse to identical, which formed as:

\begin{equation}
    \gL_{kl}=\sum_{v \in \mathcal{V}}\sum_{i\in \mathcal{N}_v} \text{KL}(\text{sim}(\mathcal{X}_v, \mathcal{X}_i)/\tau, \text{sim}(\hat{\mathcal{X}}_v, \hat{\mathcal{X}}_i)/\tau),
\end{equation}
where $\tau$ indicates the temperature parameter.

We view the strategy as knowledge distillation between raw input graph to decoding graph. The necessary distinctness between pairwise nodes get drowned in the decoding graph, results in nodes assemble each other. As a coupled structure, the encoder and decoder within GAE can be viewed as teacher and student, respectively. The masked graph (the input to the teacher) contains the distinctness that the reconstructed graph (the output of student) lacks, through the KL constraint from the teacher to the student, the decoder frees from generating vague graph. The adding of KL constraint is rather simple, compared to the standard GAEs training, the ultimate training objective combines $\gL_{rec}$ and $\gL_{kl}$ with a cofficient $\alpha$ as:
\begin{equation}
    \gL = \gL_{rec} + \alpha \times \gL_{kl}.
\end{equation}
\begin{table}[]
\centering
\caption{dataset statistics.}
\begin{tabular}{ccccc}
\hline

Dataset & Nodes & Edges & \begin{tabular}[c]{@{}c@{}}Number of \\ Classes\end{tabular} & \begin{tabular}[c]{@{}c@{}}Node feature \\ size\end{tabular}\\ \hline
Cora & 2,708 & 10,556 & 7 & 1,433 \\
Citeseer & 3,327 & 9,228 & 6 & 3,703 \\
Pubmed & 19,717 & 88,651 & 3 & 500 \\
A-Photo & 7,650 & 238,163 & 8 & 745 \\
C-CS & 18,333 & 163,788 & 15 & 6,805 \\
C-Physics & 34,493 & 495,924 & 5 & 8,415 \\ \hline
\end{tabular}
\label{tab:st}
\end{table}

\begin{table*}[thb]
\centering
\caption{Summary of the results on node classfication.}
\begin{threeparttable}
\begin{tabular}{@{}cccccccc@{}}
\toprule
Dataset               & Cora                & Citeseer            & Pubmed              & A-Photo                         & A-CS                & A-Physics           \\ \midrule
DGI                   & 82.30±0.60 & 71.80±0.70 & 76.80±0.60 & 91.61±0.22  & 92.15±0.63 & 94.51±0.52          \\
MVGRL                 &  83.50±0.40 & 73.30±0.50 & 80.10±0.70 & 91.74±0.07  & 92.11±0.12 & 95.33±0.03          \\
CCA-SSG               & 84.00±0.40 & 73.1±0.30 & 81.00±0.40 & 93.14±0.14  & 93.31±0.22 & 95.38±0.06        \\
VGAE                  & 71.50±0.40 & 65.80±0.40 & 72.10±0.50 & 92.20±0.11  & 92.11±0.09 & 75.35±0.14         \\
GraphMAE              & 84.20±0.40 & 73.40±0.40 & 81.10±0.40 & 92.98±0.35 & 93.08±0.17 & 95.30±0.12         \\
GraphMAE2             & 84.50±0.60 & 73.4±0.30 & 81.40±0.50 & -  & - & -               \\
SeeGera               & 84.30±0.40 & 73.00±0.80 & 80.40±0.40 & 92.81±0.45  &\textbf{93.84±0.11} & 95.39±0.08         \\
\textbf{Ours} & \textbf{85.35±0.35} & \textbf{73.80±0.00} & \textbf{81.70±0.10} & \textbf{93.64±0.13}    & 92.32±0.31 & 
\end{tabular}
\begin{tablenotes}
        \footnotesize
        \item[] \scriptsize "-" indicates that the corresponding performances are not reported.
    \end{tablenotes}
\end{threeparttable}
\label{table:nc}
\end{table*}

\section{Experiment}~\label{experiment}
In this section, we first introduce the fundamental experimental settings. We then evaluate the performance of our model using twelve publicly available datasets and three downstream tasks. Subsequently, we report and analyze the experimental results of the proposed method on the tasks of node classification, link prediction, and graph classification.

\subsection{Experimental Settings}
\textbf{Datasets.} The experiments for node-level and link-level tasks were conducted on six publicly accessible datasets: citation networks (Cora, Citeseer, and PubMed), co-authorship graphs (Coauthor CS and Coauthor Physics), and co-purchase graphs (Amazon Photo). The statistical details of these datasets are presented in Table~\ref{tab:st}. For training node classification and link prediction models, we adhere to the publicly available data splits for Cora, Citeseer, and PubMed, and follow the data-splitting configuration used in SeeGera~\cite{seegera} for the remaining datasets. For link prediction, 10\% of the edges constitute the test set.

For the graph classification tasks, we conducted experiments on six public datasets: IMDB-B, IMDB-M, PROTEINS, COLLAB, MUTAG, and NCI1. These datasets belong to two distinct domains: PROTEINS, MUTAG, and NCI1 are from the bioinformatics field, while the remaining datasets are derived from social networks. Each dataset consists of a collection of labeled graphs.

\textbf{Baselines.} For the node classification comparison, we selected seven SOTA methods within the SSL framework. These include three contrastive methods: DGI~\cite{dgi}, MVGRL~\cite{mvrl}, and CCA-SSG~\cite{cca}, and four generative methods: VGAE~\cite{gae}, GraphMAE, GraphMAE2~\cite{graphmae2}, and SeeGera.

For the link prediction comparison, we selected nine SOTA methods within the SSL framework. These include five contrastive methods: DGI, MVGRL, GRACE~\cite{grace}, GCA~\cite{gca}, and CCA-SSG~\cite{cca}, as well as four generative methods: CAN~\cite{can}, SIG-VAE~\cite{sig-vae}, GraphMAE, S2GAE~\cite{S2GAE}, and SeeGera.

For the graph classification comparison, we selected seven SOTA methods within the SSL framework. These include five contrastive methods: GraphCL~\cite{graphcl}, JOAO~\cite{joao}, GCC~\cite{gcc}, MVGRL, and InfoGCL~\cite{infogcl}, as well as two generative methods: GraphMAE and S2GAE~\cite{S2GAE}.

\textbf{Implementation.} In Section \uppercase\expandafter{\romannumeral 3}, we discussed the necessity of calculating the feature distance between each node and its neighbors to constrain the reconstructed representation. For C-Physics, guided by the original features, we extract the five neighboring nodes with the greatest differences for each individual node. This strategy aims to alleviate the computational burden during each training iteration.

Regarding the evaluation protocol, we adhere to the experimental framework outlined in ~\cite{graphmae}. During the training stages, a GNN encoder is trained using \model without supervision. Following this, the parameters of the encoder are fixed, enabling the generation of node embeddings for all nodes. The resultant node embeddings, denoted as $Z$, are utilized for tasks at the node, link, and graph levels. In the case of link prediction, the adjacency matrix is obtained by computing \( \sigma (ZZ^{T}) \). For the graph-level task, a graph pooling readout function is applied to $Z$ to obtain the graph-level representations.

\begin{table*}[]
\centering
\caption{Experimental results for link prediction.}
\label{tab:lp}
\begin{threeparttable}
\begin{tabular}{@{}c|cccccccc@{}}
\toprule
Metrics               & Method   & Cora                & Citeseer            & Pubmed              & A-Photo                      & A-CS                  & A-Physics             \\ \midrule
\multirow{10}{*}{AP}  & DGI      & 93.60±1.14          & 96.18±0.68          & 95.65±0.26          & 81.01±0.47                  & 92.79±0.31          & 92.10±0.29          \\
                      & MVGRL    & 92.95±0.82          & 89.37±4.55          & 95.53±0.30          & 63.43±2.02           
                      & 89.14±0.93          & -                   \\
                      & GRACE    & 82.36±0.24          & 86.92±1.11          & 93.26±1.20          & 81.18 ± 0.37         
                      & 83.90±2.20          & 82.20±1.06          \\
                      & GCA      & 80.87±4.11          & 81.93±1.76          & 93.31±0.75          & 65.17±10.11                   & 83.24±1.16          & 82.80±4.46          \\
                      & CCA-SSG  & 93.74±1.15          & 95.06±0.91        & 95.97±0.23          & 67.99±1.60          
                      & 96.40±0.30          & 96.26±0.10          \\
                      & CAN      & 94.49±0.60          & 95.49±0.61          & -                   & 96.68±0.30              
                      & -                   & -                   \\
                      & SIG-VAE  & 94.79±0.71          & 94.21±0.53          & 85.02±0.49          & 94.53±0.93              
                      &94.93±0.37          & 98.85±0.12          \\
                      & GraphMAE & 89.52±0.01          & 74.50±0.04          & 87.92±0.01          & 77.18± 0.02              
                      &83.58±0.01          & 86.44±0.03          \\
                      & SeeGERA  & 95.92±0.68          & 97.33±0.46          & \textbf{97.87±0.20}          & \textbf{98.48±0.06} 
                      & \textbf{98.53±0.18}          & \textbf{99.18±0.04} \\ \cmidrule(l){2-8} 
                      & \model     & \textbf{98.10±0.12} & \textbf{99.27±0.12} & 97.73±0.02 & 96.60±0.05            
                      & 97.15±0.01 & 95.28±0.09         \\ \midrule
\multirow{11}{*}{AUC} & DGI      & 93.88±1.00          & 95.98±0.72          & 96.30±0.20          & 80.95±0.39                   & 93.81±0.20          & 93.51±0.22       \\
                      & MVGRL    & 93.33±0.68          & 88.66±5.27          & 95.89±0.22          & 69.58±2.04          
                      & 91.45±0.67          & -                   \\
                      & GRACE    & 82.67±0.27          & 87.74±0.96          & 94.09±0.92          & 81.72±0.31          
                      & 85.26±2.07          & 83.48±0.96          \\
                      & GCA      & 81.46±4.86          & 84.81±1.25          & 94.20±0.59          & 70.02±9.66            
                      & 84.35±1.13          & 85.24±5.41          \\
                      & CCA-SSG  & 93.88±0.95          & 94.69±0.95          & 96.63±0.15          & 73.98±1.31     
                      & 96.80±0.16          & 96.74±0.05          \\
                      & CAN      & 93.67±0.62          & 94.56±0.68          & -                   & 97.00±0.28       
                      & -                   & -                   \\
                      & SIG-VAE  & 94.10±0.68          & 92.88±0.74          & 85.89±0.54          & 94.98±0.86               
                      & 95.26±0.36          & 98.76±0.23          \\
                      & GraphMAE & 90.70±0.01          & 70.55±0.05          & 69.12±0.01          & 77.42±0.02            
                      & 91.47±0.01          & 87.61±0.02          \\
                      & S2GAE    & 93.52±0.23          & 93.29±0.49          &\textbf{98.45±0.03}          & -                                  & -                   & -                   \\
                      & SeeGERA  & 95.50 ± 0.71        & 97.04±0.47          & 97.87±0.20       & \textbf{98.64±0.05}  
                      & \textbf{98.42±0.13}          & \textbf{99.03±0.05} \\ \cmidrule(l){2-8} 
                      & \model     & \textbf{98.02±0.12} & \textbf{99.22±0.12} & 97.85±0.02 & 97.11±0.03                  
                      & 97.68±0.08 & 96.3±0.06 \\ \bottomrule
\end{tabular}
\begin{tablenotes}
        \footnotesize
        \item[] \scriptsize "-" indicates that unavailable code or out-of-memory.
    \end{tablenotes}
\end{threeparttable}
\label{table:lp}
\end{table*}

\begin{table*}[htb]
\centering
\caption{experimental results for graph classification. }
\label{tab:gc}
\begin{threeparttable}

\begin{tabular}{@{}ccccccccc@{}}
\toprule
\multicolumn{1}{l}{}           & Dataset   & \multicolumn{1}{l}{IMDB-B} & \multicolumn{1}{l}{IMDB\-M} & \multicolumn{1}{l}{PROTEINS} & \multicolumn{1}{l}{COLLAB} & \multicolumn{1}{l}{MUTAG}  & NCI \\ 
\midrule
\multirow{5}{*}{Contrastive}   & GraphCL   & 71.14±0.44                  & 48.58±0.67                       & 74.39±0.45                   & 71.36±1.15                      & 86.80±1.34                           & 77.87±0.41              \\
                               & JOAO      & 70.21±3.08                  & 49.20±0.77                       & 74.55±0.41                   & 69.50±0.36                      & 87.35±1.02                            & 78.07±0.47              \\
                               & GCC       & 72.0                        & 49.4                             & -                            & 78.9                            & -                                      & -                       \\
                               & MVGRL     & 74.20±0.70                  & 51.20±0.50                       & -                            & -                               & 89.70±1.10                               & -                       \\
                               & InfoGCL   & 75.10±0.90                  & 51.40±0.80                       & -                            & 80.00±1.30                      & \textbf{91.20±1.30}                                  & 80.20±0.60              \\ \midrule
\multirow{3}{*}{Generative}    & GraphMAE  & 75.52±0.66                  & 51.63±0.52                       & \underline{75.30±0.39}                   & 80.32±0.46                      & 88.19±1.26                               & \underline{80.40±0.30}              \\
                               & S2GAE     & \underline{75.76±0.62}          & \underline{51.79±0.36}               & \textbf{76.37±0.43}        & \underline{81.02±0.53}              & 88.26±0.76                           & \textbf{80.80±0.24}   \\
                               & \model & \textbf{77.8±1.47}         & \textbf{52.2±2.7}              & 74.04±4.47             & \textbf{81.44±1.59}           & \underline{88.83±6.18}                      & 75.91±2.1     \\ \bottomrule
\end{tabular}
\begin{tablenotes}
        \footnotesize
        \item[] \scriptsize "-" indicates that unavailable code or out-of-memory.
    \end{tablenotes}
\end{threeparttable}
\end{table*}

\textbf{Evaluation.} 
We selected four widely-used and complementary metrics for different tasks: accuracy (Acc), area under the characteristic curve (AUC) and the average precision (AP). To optimize the training process, we applied the Adam optimizer with a learning rate ranging from 0.0001 to 0.01. The temperature parameter \(\tau\) and the coefficient \(\alpha\) were tuned from 1 to 10 and 0.5 to 20, respectively. The results are reported as the mean ± standard deviation, calculated over ten experimental repetitions.

\subsection{Comparison on Node Classification}
We use Acc as the metric to evaluate node classification. The node classification results are displayed in Table~\ref{table:nc}. From these results, we observe that \model outperforms the selected contrastive and generative methods in most cases. Furthermore, these results demonstrate that constraining the pairwise similarity between nodes during the reconstruction process can aid in effectively reconstructing the original nodes, thereby assisting in generating superior embedding representations and improving the precision of node classification outcomes. It is noteworthy that the performance enhancement result for C-Physics is not satisfactory. We restricted and extracted the neighboring nodes with the maximum feature distance for C-Physics. Our analysis indicates that solely preserving the differences in the reconstruction process does not necessarily result in enhanced reconstruction outcomes. It is essential to note that instead of preserving both similarity and dissimilarity among nodes during the reconstruction process, the methodology should place emphasis on moderately retaining differences grounded in similarity.

\subsection{Comparison on Link Prediction}
We use AP and AUC as metrics to evaluate link prediction. Table~\ref{table:lp} reports the link prediction results. In the experimental settings, we have delineated the evaluation methodology for link prediction. Although our emphasis lies on the effective reconstruction of node features, the performance of \model in link prediction remains competitive with methods specifically designed for topological relationships, such as Seegera. However, compared to the performance in node classification, our strategy did not achieve superior outcomes across multiple datasets in the link prediction task. This can be attributed to the relatively suboptimal performance of the feature reconstruction model in link prediction. Our approach is founded on GraphMAE, and its efficacy can be directly substantiated through a contrastive analysis with GraphMAE. Compared to GraphMAE, we have achieved enhanced prediction results across all datasets. This indicates that generating embeddings by upholding the differences among nodes during the reconstruction phase has positively influenced link prediction performance.

\subsection{Comparison on Graph Classification}
We use the F1 score as the metric to evaluate the task of graph classification. Our evaluation was further broadened to incorporate graph classification tasks, with the results detailed in Table~\ref{tab:gc}. When compared to generative approaches, our proposed method delivered superior outcomes across diverse datasets. This provides further validation of our method's efficacy in optimizing reconstruction results. 
The evidence from Table~\ref{tab:gc} distinctly demonstrates the stronger performance of our methodology on social network datasets compared to those from the bioinformatics field. We interpret this phenomenon as reflective of the significance of preserving the distinctiveness of nodes within social networks. Specifically, the presence of highly distinctive node features within social networks contributes considerably to enabling more precise network differentiation.

\subsection{Comparison on Pairwise Similarity}
To ascertain whether our proposed methodology can assist in the reconstruction and generation of more distinct nodes, we independently calculated the average neighbor feature similarity for nodes, considering scenarios both with and without the strategy applied. Herein, $v_1$ corresponds to the vanilla masked GAE, whereas $v_2$ symbolizes the version supplemented with KL constraint. Concurrently, \(Encoder\) and \(Decoder\) signify the generated latent node embedding and the reconstructed node embedding, respectively. As displayed in the table~\ref{table:pair}, our strategy successfully reconstructed nodes with enhanced distinction on all three datasets, while also generating node embeddings of elevated distinctiveness.
\begin{table}[]
\centering
\caption{mean pairwise similarity on 3 datasets.}
\begin{tabular}{cccc}
\hline
 & Cora & Citeseer & Pubmed \\ \hline
Raw & 0.1779 & 0.2030 & 0.2657 \\
$Encoder_{v_1}$ & 0.9648 & 0.9489 & 0.9194 \\
$Encoder_{v_2}$ & 0.9502 & 0.9117 & 0.8672 \\
$Decoder_{v_1}$ & 0.8705 & 0.8696 & 0.8095 \\
$Decoder_{v_2}$ & 0.8560 & 0.8578 & 0.7923 \\ \hline
\end{tabular}
\label{table:pair}
\end{table}

\section{Conclusion}
In this study, we focus on improving the reconstruction of masked graph autoencoders. We conduct an analysis showing that relying solely on MSE or SCE as the reconstruction criterion may result in the loss of necessary distinctness between nodes. To address this, we develop a simple yet effective strategy, i.e., the KL divergence constraint on the pairwise nodes similarity between the raw graph and the reconstructed graph. We conduct extensive experiments, validating the effectiveness and generality of the strategy.

\bibliographystyle{splncs04}
\bibliography{icdm}
\end{document}